\documentclass[letterpaper, 10 pt, conference]{ieeeconf}  

\IEEEoverridecommandlockouts                              

\usepackage{graphicx} 
\usepackage{xcolor}
\usepackage{amsmath}
\usepackage{amssymb}
\usepackage{amsfonts}
\usepackage{cite}
\usepackage{amsmath}
\usepackage{amssymb}
\usepackage{amsfonts}
\usepackage{mathrsfs}
\usepackage{booktabs}
\usepackage{multirow}
\usepackage{colortbl}

\usepackage[colorlinks=true, linkcolor=blue, citecolor=blue, urlcolor=blue]{hyperref}

\title{DinoLink: A Token-Centric Representation Compression Framework for Bandwidth-Constrained Collaborative V2X Perception
}

\author{Tianle Zhu$^{1,\dagger}$, Haohua Que$^{1,\dagger}$, Handong Yao$^{1,*}$, Hongyi Xu$^{2}$, and Zhipeng Bao$^{1}$
\thanks{$^{\dagger}$These authors contributed equally. $^{*}$Corresponding author: Handong Yao.}%
\thanks{$^{1}$University of Georgia, Athens, GA, USA. $^{2}$University of the Arts London, London, UK. {\tt\small Handong.Yao@uga.edu}}
}

\begin{document}

\maketitle
\thispagestyle{empty}
\pagestyle{empty}

\begin{abstract}
High-precision remote perception is often hindered by the severe bandwidth constraints of Vehicle-to-Everything (V2X) networks. We propose \textit{DinoLink}, a token-centric compression framework that replaces raw pixel streaming with discrete semantic communication for vehicle-cloud collaborative inference. DinoLink employs a dual-sparsity architecture: a saliency-aware selector prunes redundant background tokens, while a Residual Vector Quantization (RVQ) module collapses features into compact codebook indices. By transmitting only lightweight indices and positional priors, DinoLink achieves a $139\times$ bitrate reduction compared to uncompressed transmission while maintaining a competitive 32.8\% mAP on the nuScenes dataset. Deployment simulations further demonstrate a $34.5\times$ acceleration in narrow-band environments, such as LoRa. Our results substantiate DinoLink as a robust, bandwidth-efficient frontend for high-fidelity remote perception in constrained V2X scenarios. The code is publicly available at \url{https://github.com/UGA-MOBILITY-LAB/dino_link}.
\end{abstract}


\section{Introduction}
\label{sec:intro}
The deployment of foundation models, such as DINOv2~\cite{oquab2023dinov2} and DETR~\cite{carion2020end,zhu2020deformable}, has ushered in a new era of high-precision perception for autonomous driving.
However, executing these massive architectures exclusively on vehicle-edge computing platforms hits a rigid compute and thermal wall~\cite{chen2019deep}.  Consequently, the industry is increasingly pivoting towards Vehicle-Cloud Collaborative Inference: offloading heavy perception backends to the cloud while leaving lightweight frontends on the vehicle~\cite{kang2017neurosurgeon}.
Yet, this promising paradigm is currently paralyzed by the strict bandwidth constraints and unpredictable latency of V2X networks~\cite{karagiannis2011vehicular}.

Existing collaborative perception schemes suffer from a fundamental misalignment between communication protocols and machine semantics.
Transmitting highly compressed images (e.g., JPEG or H.264) relies on codecs strictly optimized for the Human Visual System (HVS)~\cite{wallace1991jpeg,wiegand2003overview}.
These codecs aggressively truncate high-frequency spatial components and color chroma to save bandwidth.
While visually imperceptible to human drivers, this irreversible compression destroys the fine-grained semantic priors and localized gradient information essential for machine vision backends~\cite{dodge2016understanding,hendrycks2019benchmarking}.
Conversely, direct transmission of intermediate continuous neural tensors (e.g., Float32 feature maps) bypasses the HVS bottleneck, uncompressed feature maps often exceed the size of the original images~\cite{kang2017neurosurgeon, teerapittayanon2017distributed}.
This inflates the per-frame payload to several megabytes, completely overwhelming the inherently volatile and constrained bandwidth of real-world V2X networks, introducing fatal transmission latency, and rendering real-time collaborative driving practically unfeasible~\cite{karagiannis2011vehicular, LI2024100136}.

To resolve this bottleneck, we propose \emph{DinoLink}, a token-centric transmission framework explicitly designed for collaborative machine perception.
DinoLink transitions from pixel-level to \emph{discrete semantic-centric} communication via a funnel-like dual-sparsity architecture.
Operating on the vehicle edge, DinoLink employs a frozen DINOv2 backbone to extract semantic tokens. To achieve extreme bandwidth efficiency, the first stage of our funnel introduces a Saliency-Aware Token Selector, which functions as a spatial filter.
Using a Top-$K$ masking strategy, we aggressively discard redundant background tokens (e.g., sky, empty roads), achieving spatial sparsity~\cite{rao2021dynamicvit, ryoo2021tokenlearner}.
Crucially, rather than quantizing the entire dense feature map, the surviving critical tokens are channeled through a Residual Vector Quantization (RVQ) module~\cite{zeghidour2021soundstream,defossez2022high}.
By repurposing RVQ as an ultra-efficient communication protocol, we collapse continuous features into a compact sequence of discrete codebook indices (bit-level sparsity)~\cite{van2017neural}.
The cloud server receives only these lightweight integer indices and their positional coordinates, seamlessly reconstructing the high-dimensional semantic token space via a token decoder~\cite{razavi2019generating}, which is subsequently processed by an off-the-shelf DETR backend~\cite{carion2020end,zhu2020deformable}.

Overall, our contributions are summarized as follows. First, we propose \emph{DinoLink}, a highly efficient vehicle-cloud V2X collaborative perception framework that successfully bridges DINOv2 and DETR over strictly constrained V2X networks via discrete semantic token transmission. Second, we design a funnel-like dual-sparsity pipeline that synergistically integrates Saliency-Aware Top-$K$ selection (spatial filtering) with subsequent Residual Vector Quantization (bit-level compression), effectively repurposing generative tools for autonomous driving communication. Finally, we extensively evaluate the system using both the nuScenes dataset~\cite{caesar2020nuscenes} and practical on-vehicle deployment. The experimental results reveal an unprecedented bandwidth-accuracy trade-off, verifying that \emph{DinoLink} can preserve critical machine semantics even when subjected to extreme payload compression in real-world driving scenarios.

\section{Related Work}
\label{sec:related_work}

\subsection{Collaborative Perception in Autonomous Driving}
The increasing demand for robust autonomous driving has driven a paradigm shift from single-agent perception to Vehicle-to-Everything (V2X) collaborative perception~\cite{karagiannis2011vehicular,chen2017vehicle,xiang2009wireless}.
By sharing sensory data or processed features between vehicles (V2V) or between vehicles and cloud/edge servers (V2I/V2N), collaborative systems fundamentally mitigate occlusion and extend the perception range~\cite{wang2020v2vnet,yu2022dair,xu2022opv2v,xu2022v2x}.
To address the compute-thermal constraints of vehicle edge devices, ``split computing" has been actively explored \cite{kang2017neurosurgeon,teerapittayanon2017distributed,zhu2025wireless,li2025wireless}. Cloud-side backends have also been adopted in roadside perception systems for real-time vehicle tracking and control\cite{10586907}.
In this paradigm, lightweight feature extractors operate on the vehicle, while heavy perception backends are offloaded to the cloud. However, existing split-computing frameworks typically transmit dense intermediate feature maps (e.g., continuous Float32 tensors).
As spatial resolution and channel dimensions grow in modern architectures, these feature payloads frequently exceed the size of raw images, imposing severe transmission latency over bandwidth-constrained and volatile wireless networks~\cite{chen2019deep,hendrycks2019benchmarking}.
Unlike these approaches, DinoLink completely abandons dense continuous feature transmission, proposing a discrete semantic-centric pipeline to strictly bound the communication payload.

\subsection{Vision Foundation Models for Transportation}
Vision foundation models, primarily based on the Vision Transformer (ViT) architecture~\cite{dosovitskiy2020image,liu2021swin}, have demonstrated unprecedented generalization and semantic understanding in transportation contexts~\cite{WANG2025100208}. Models like DINOv2~\cite{oquab2023dinov2,caron2021emerging,he2022masked,zhou2021ibot} utilize self-supervised learning to produce powerful patch-level semantic tokens, which serve as excellent priors for downstream tasks without task-specific fine-tuning.
Concurrently, DETR-based architectures~\cite{carion2020end,zhu2020deformable} have redefined object detection by streamlining the pipeline with end-to-end set prediction.
While foundation-model tokens and DETR-style set prediction are strong building blocks for modern driving perception~\cite{chitta2022transfuser,li2024bevformer}, deploying such massive models directly on edge devices is often prohibitive. DinoLink bridges this gap by enabling foundation-model-level perception over the cloud, ensuring that the critical semantic tokens extracted by DINOv2 on the edge can be efficiently compressed, transmitted, and perfectly interpreted by the DETR backend on the server.

\subsection{Task-Oriented Feature Compression}
Traditional image and video codecs, such as JPEG and HEVC, are heavily optimized for the Human Visual System (HVS)~\cite{wallace1991jpeg,wiegand2003overview}.
While they achieve high compression ratios for human consumption, they irreversibly discard high-frequency spatial details and gradients that are vital for machine vision tasks, leading to severe performance drops in downstream object detection~\cite{dodge2016understanding,hendrycks2019benchmarking}.
Recent advancements in neural compression~\cite{defossez2022high,theis2017lossy,balle2016end,NEURIPS2018_53edebc5,10.5555/3305890.3305983,FullResolutionImageCompression,cheng2020image,mentzer2020high} have explored task-oriented feature compression, aiming to transmit features rather than pixels. Among discrete representation learning techniques, Vector Quantization (VQ)~\cite{van2017neural} maps continuous vectors to discrete codebooks~\cite{esser2020taming}.
To overcome the codebook collapse and capacity limits of standard VQ, Residual Vector Quantization (RVQ) was introduced, partitioning the quantization process into a coarse-to-fine multi-stage residual approximation~\cite{razavi2019generating}.
RVQ has been remarkably successful in high-fidelity autoregressive image generation and audio compression~\cite{zeghidour2021soundstream}.
In this work, we repurpose the generative RVQ technique as an ultra-low bandwidth V2X communication protocol. By synergistically integrating it with a spatial Top-$K$ selector, DinoLink establishes an extreme dual-sparsity framework explicitly designed for bandwidth-constrained collaborative perception.

\section{Problem Formulation}
Consider a vehicle-to-server perception system, where an edge vehicle captures observations and transmits compact representations to a remote server over a bandwidth-limited V2X link.
Let $I^t$ denote the raw observation at time $t$ (e.g., an RGB image), and let $Y^t$ denote the corresponding ground-truth output of a downstream task (e.g., 3D object detection labels).
Due to communication constraints, the vehicle cannot transmit $I^t$ in full resolution and frequency.
Instead, the vehicle sends a compact representation $Z^t$ to the server, from which the server produces the prediction $\hat{Y}^t$.

We formulate the objective as maximizing downstream task performance under a per-frame communication budget:
\begin{equation}
	\begin{aligned}
		\max_{\theta,\phi}\ \ & \sum_{t=1}^{T} g\!\left(\hat{Y}^{t}, Y^{t}\right) \\
		\text{s.t.}\ \ & \hat{Y}^{t} = f_{\theta}\!\left(Z^{t}\right), \\
		& Z^{t} = Q_{\phi}\!\left(I^{t}\right), \\
		& R\!\left(Z^{t}\right) \le B,
	\end{aligned}
	\label{eq:problem_formulation}
\end{equation}
where $f_{\theta}(\cdot)$ denotes the server-side downstream model with trainable parameters $\theta$,
$Q_{\phi}(\cdot)$ denotes the edge-side representation encoder/quantizer with trainable parameters $\phi$,
$R(\cdot)$ measures the number of transmitted bits (or bytes) for one frame,
and $B$ is the communication budget.
In DinoLink, $Z^t$ is an explicitly packetized representation consisting of discrete RVQ indices and token positions, yielding a strictly bounded payload controlled by $(K, M, |\mathcal{E}|)$.

\section{METHODOLOGY}
\begin{figure*}[t]
  \centering
  \includegraphics[width=0.9\textwidth]{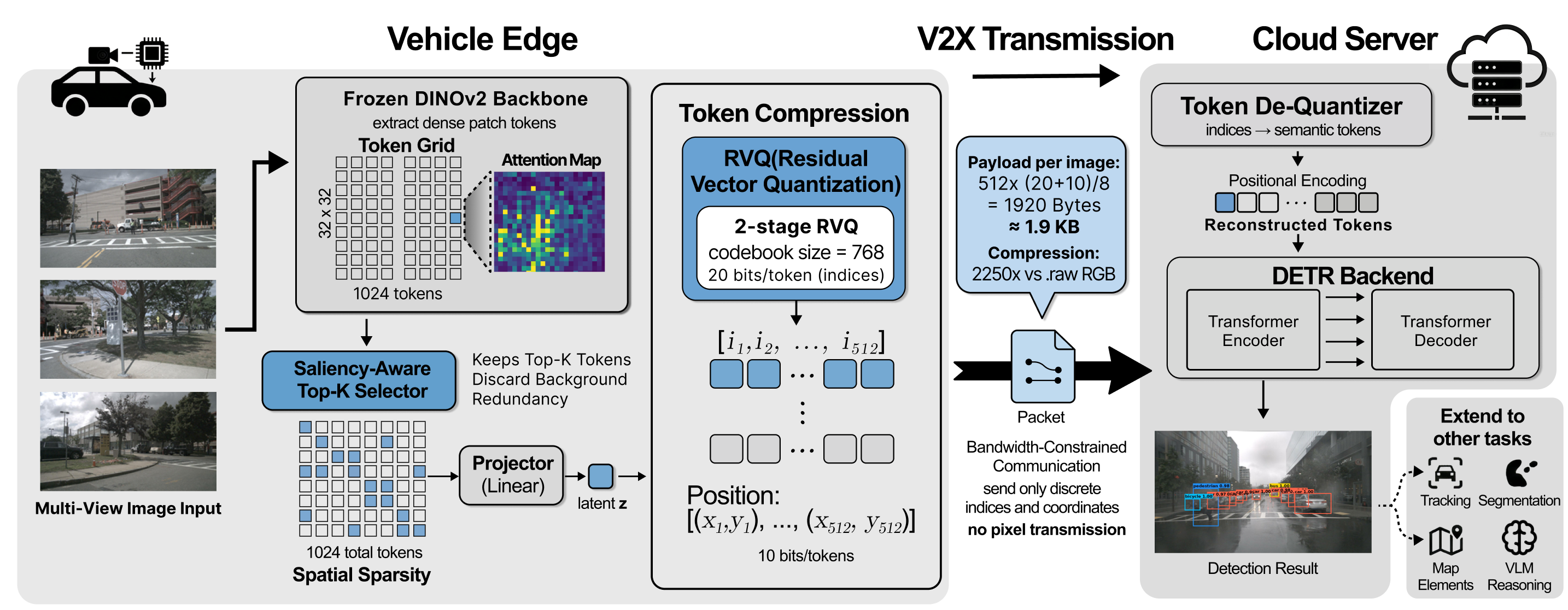}
  \caption{Overview of DinoLink. The edge vehicle extracts dense DINOv2 tokens, selects Top-$K$ salient tokens (top 50\% token selected case), and compresses them via RVQ ($C=768$, codebook size per stage) into discrete indices with lightweight positional priors for V2X transmission. The cloud server decodes the tokens and feeds them into an off-the-shelf Transformer backend (e.g., DETR) for downstream perception under a strict bandwidth budget.}
  \label{fig:overview}
  \vspace{-20pt}
\end{figure*}

\subsection{Overall Architecture}
\label{sec:overview}

As shown in Fig.~\ref{fig:overview}, DinoLink is a token-centric transmission framework that replaces pixel streaming with discrete semantic communication.
Given an observation $I^t$ on the vehicle, a self-supervised visual encoder first extracts dense patch-level tokens.
A saliency-aware Top-$K$ selector then retains a small subset of informative tokens, forming a sparse semantic representation and removing redundant background content (spatial sparsity).
The selected tokens are next compressed by Residual Vector Quantization (RVQ) on the vehicle, which maps continuous embeddings into discrete codebook indices (bit-level sparsity).
Only the quantized indices, together with lightweight positional priors, are transmitted over the bandwidth-limited V2X link.

On the server, a lightweight token decoder reconstructs continuous token embeddings from the received indices and positions.
The decoded tokens are subsequently consumed by query-driven downstream perception models deployed on the server (e.g., Transformer-based detection and 3D perception backbones).
This design decouples the compression interface from any specific downstream architecture: through a lightweight adapter, rendering DinoLink a versatile, plug-and-play frontend for diverse downstream perception tasks.

\emph{Trainable components:}
Unless otherwise stated, DinoLink keeps the DINOv2 encoder frozen and trains the projector, RVQ codebooks, token decoder, and (optionally) the downstream head.
Top-$K$ selection is performed deterministically from DINOv2 attention responses and does not require gradients.
The training objective jointly encourages (i) reconstruction-faithful token decoding, (ii) stable quantization via standard codebook/commitment regularization, and (iii) task-discriminative representations via a downstream loss.

\begin{figure}[t]
  \centering
  \includegraphics[width=0.85\columnwidth]{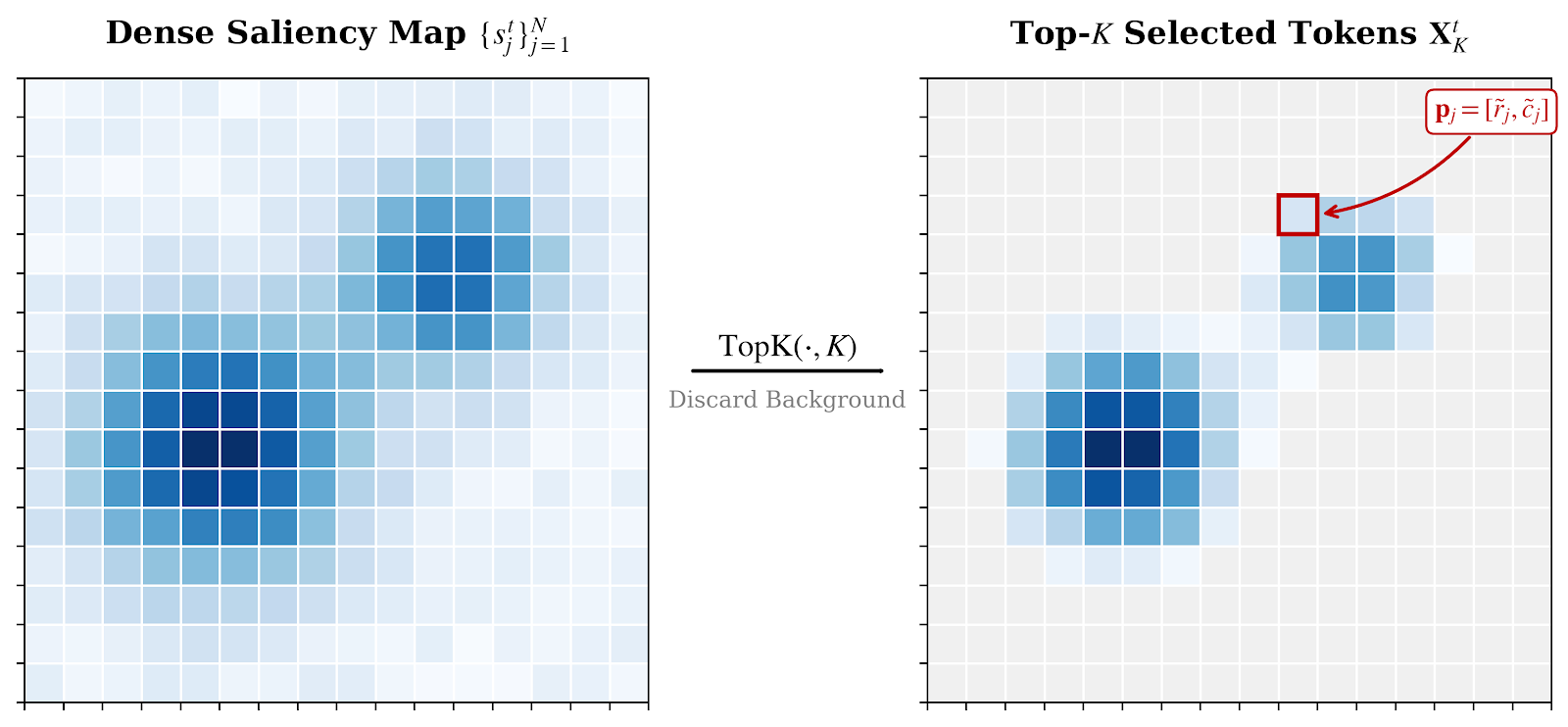}
  \caption{\textbf{Saliency-Aware Top-$K$ Token Selection.} (Left) Dense saliency map $s_j^t$ derived from DINOv2 self-attention. (Right) Sparse token set $\mathbf{X}_K^t$ after filtering redundant background. Normalized 2D positions $\mathbf{p}_j$ are retained for server-side spatial reconstruction.}
\label{fig:topk_selection}
\vspace{-15pt}
\end{figure}

\subsection{DINOv2 Token Extraction and Top-$K$ Selection (Fig.~\ref{fig:topk_selection})}
\label{sec:token_select}

We adopt DINOv2~\cite{oquab2023dinov2} as a self-supervised visual encoder to extract patch-level tokens.
Given an input image $I^t$, DINOv2 outputs a dense token set
$\mathbf{X}^t=\{\mathbf{x}_j^t\}_{j=1}^{N}$ (e.g., $N=H_pW_p$ tokens on a patch grid),
along with self-attention maps that reflect the relative saliency of spatial regions.
We compute a scalar saliency score $s_j^t$ for each token (e.g., by aggregating attention weights over heads/layers),
and select the Top-$K$ tokens:
\begin{equation}
	\mathcal{S}^t = \mathrm{TopK}\!\left(\{s_j^t\}_{j=1}^{N}, K\right),\quad
	\mathbf{X}_{K}^t=\{\mathbf{x}_j^t\mid j\in\mathcal{S}^t\}.
\end{equation}
This operation retains salient foreground content and discards redundant background tokens, resulting in spatial sparsity.

For each selected token, we also record its 2D position on the patch grid.
Let $j\mapsto (r_j,c_j)$ denote the row/column coordinates.
We normalize the position to $\mathbf{p}_j=[\tilde{r}_j,\tilde{c}_j]\in[-1,1]^2$ and include it as lightweight side information for server-side decoding.
Importantly, DinoLink does not require access to downstream queries for token selection; instead, it preserves semantically salient regions that are typically informative for query-driven downstream transformers.

\subsection{Token Quantization with Residual Vector Quantization}
\label{sec:rvq}
To reduce communication cost, we quantize the selected token embeddings into discrete codes using Residual Vector Quantization (RVQ).
A lightweight projector first maps DINOv2 tokens to a quantization-friendly latent space:
\begin{equation}
	\mathbf{z}_j^t = h_{\phi}\!\left(\mathbf{x}_j^t\right), \quad j\in\mathcal{S}^t,
\end{equation}
where $h_{\phi}$ is a small trainable MLP (or linear layer).

RVQ progressively approximates each latent vector via $M$ codebooks.
Given $\mathbf{z}\in\mathbb{R}^d$, we set $\mathbf{r}^{(0)}=\mathbf{z}$ and for stage $m=1,\ldots,M$:
\begin{equation}
\begin{aligned}
\mathbf{r}^{(0)} &= \mathbf{z},\\
i^{(m)} &= \arg\min_{k}\left\|\mathbf{r}^{(m-1)}-\mathbf{e}^{(m)}_{k}\right\|_{2}^{2},\\
\mathbf{q}^{(m)} &= \mathbf{e}^{(m)}_{i^{(m)}},\\
\mathbf{r}^{(m)} &= \mathbf{r}^{(m-1)}-\mathbf{q}^{(m)}, \qquad m=1,\ldots,M.
\end{aligned}
\end{equation}
and obtain the quantized vector
\begin{equation}
	\mathbf{z}_{q}=\sum_{m=1}^{M}\mathbf{q}^{(m)}.
\end{equation}
The transmitted code for one token is the index tuple $\mathbf{i}=[i^{(1)},\ldots,i^{(M)}]$.
With $K$ selected tokens, the per-frame payload is strictly bounded by
$K\cdot M\cdot \lceil\log_2|\mathcal{E}|\rceil$ bits for indices plus a small overhead for positions.

\emph{Transmitted packet:}
For each frame, DinoLink transmits
\begin{equation}
	Z^t = \left\{ \left(\mathbf{i}_j^t,\mathbf{p}_j^t\right)\right\}_{j\in\mathcal{S}^t},
\end{equation}
i.e., discrete RVQ indices and normalized 2D positions for the selected tokens.
No pixels or dense Float32 feature maps are sent.

\subsection{Token Decoder and Downstream Integration}
\label{sec:decoder}
Upon receiving $Z^t$, the server first reconstructs quantized latents by codebook lookup:
\begin{equation}
	\mathbf{z}_{q,j}^t = \sum_{m=1}^{M} \mathbf{e}^{(m)}_{i_{j}^{(m)}},\quad j\in\mathcal{S}^t.
\end{equation}
A lightweight token decoder $d_{\phi}$ then maps each $\left[\mathbf{z}_{q,j}^t,\mathbf{p}_j^t\right]$ to a decoded token $\hat{\mathbf{x}}_j^t$:
\begin{equation}
	\hat{\mathbf{x}}_j^t
	= \mathrm{LN}\!\left(
	\mathbf{W}_3\, \sigma\!\left(
		\mathbf{W}_2\, \sigma\!\left(
			\mathbf{W}_1 \left[\mathbf{z}_{q,j}^t, \mathbf{p}_j^t\right] + \mathbf{b}_1
		\right) + \mathbf{b}_2
	\right) + \mathbf{b}_3
	\right),
\end{equation}
where $\sigma(\cdot)$ is GELU, $[\cdot,\cdot]$ denotes concatenation, and $\mathrm{LN}(\cdot)$ is LayerNorm.
This yields a sparse decoded token set $\hat{\mathbf{X}}_{K}^t=\{\hat{\mathbf{x}}_j^t\}_{j\in\mathcal{S}^t}$.

\emph{Adapter to downstream models:}
Many Transformer-based perception models can consume token sets with positional encoding.
When a downstream backbone expects a dense grid, we scatter the sparse tokens back to their patch-grid locations and fill missing tokens with zeros (or a learned mask token), while preserving the same positional encoding convention.
When a downstream backbone supports sparse token inputs, we directly feed $\hat{\mathbf{X}}_{K}^t$ with positions $\{\mathbf{p}_j^t\}$.
In both cases, the downstream model operates purely on the server and can be replaced without modifying the edge-side transmission protocol.

\subsection{Training Objective}
\label{sec:loss}

DinoLink is trained to (i) reconstruct selected semantic tokens faithfully after quantization and transmission, and (ii) support downstream task learning under the same compressed interface.
Let $\mathbf{x}_j^t$ denote the original DINOv2 token and $\hat{\mathbf{x}}_j^t$ the decoded token for $j\in\mathcal{S}^t$.
We define the token reconstruction loss over the selected token set:
\begin{equation}
	\mathcal{L}_{t}
	= \frac{1}{|\mathcal{S}^t|}\sum_{j\in\mathcal{S}^t}
	\left(
	\lambda_{2}\, \|\hat{\mathbf{x}}_j^t - \mathbf{x}_j^t\|_2^2
	+ \lambda_{l}\, \mathcal{L}_{l}(\hat{\mathbf{x}}_j^t, \mathbf{x}_j^t)
	\right),
\end{equation}
where $\mathcal{L}_l(\cdot,\cdot)$ is the Logit-Laplace negative log-likelihood term that encourages robustness to heavy-tailed reconstruction residuals.

Following the standard VQ formulation, the quantization regularization includes a codebook loss and an encoder commitment loss:
\begin{equation}
	\mathcal{L}_{q}
	= \|\mathbf{z}_q - \mathrm{sg}(\mathbf{z}_e)\|_2^2
	+ \beta \|\mathbf{z}_e - \mathrm{sg}(\mathbf{z}_q)\|_2^2,
\end{equation}
where $\mathrm{sg}(\cdot)$ denotes stop-gradient, $\mathbf{z}_e$ is the projected latent, and $\mathbf{z}_q$ is the quantized latent.

If training with a downstream perception model, we include the task-specific loss $\mathcal{L}_d$ (e.g., DETR classification and box regression losses for detection).
The overall objective is
\begin{equation}
	\mathcal{L}
	= \lambda_{t}\, \mathcal{L}_{t}
	+ \mathcal{L}_{q}
	+ \mathcal{L}_{d}.
\end{equation}
In practice, we keep the encoder frozen and optimize the parameters of the projector, RVQ codebooks, token decoder, and the downstream head (when applicable), ensuring that the transmitted discrete codes remain bandwidth-efficient while preserving task-relevant semantics.

\section{Experiment and Results}

\subsection{Experiment Setups}

\subsubsection{Downstream Task}
As the downstream task, we adopt DETR~\cite{carion2020end} as our 2D object detection framework. DETR formulates object detection as a set prediction problem and removes the need for heuristic components such as anchor design and non-maximum suppression (NMS). The architecture consists of a backbone network for feature extraction followed by a transformer encoder–decoder that models global context through self-attention and cross-attention. Object queries in the decoder directly predict a fixed-size set of bounding boxes and class labels, which are optimized via bipartite matching using the Hungarian algorithm.
To evaluate the effectiveness of our representation, we replace the original convolutional backbone in DETR with a DINO-based Vision Transformer backbone. The DINO backbone provides semantically rich and globally consistent features, which are particularly beneficial for transformer-based detection architectures. All other components of DETR remain unchanged to ensure a controlled comparison.

\subsubsection{Dataset}
We evaluate our method on the nuScenes dataset \cite{caesar2020nuscenes}, originally designed for 3D object detection in autonomous driving scenarios, which provides 3D bounding box annotations defined in the global coordinate frame together with calibrated multi-camera sensor parameters. To construct a 2D detection benchmark, we project the annotated 3D bounding boxes onto the image plane using the official nuScenes development toolkit (nuscenes-devkit). Concretely, 3D box corners are transformed from the global frame to each camera coordinate frame via the provided extrinsic calibration, followed by perspective projection using the intrinsic camera matrix; the final 2D bounding boxes are obtained by computing the tight axis-aligned bounding rectangles enclosing the projected corner points, and the annotations are formatted in COCO style for compatibility with standard 2D detection frameworks. The benchmark follows the standard nuScenes detection taxonomy with 10 object categories. For computational efficiency, we randomly sample 5,000 images from the dataset, where camera views are randomly selected from the six synchronized surround-view cameras using a fixed random seed of 42 to ensure reproducibility. All methods are trained and evaluated on the identical data subset, category definitions, and annotation settings to guarantee a controlled and fair comparison, without introducing any external data or additional supervision.

\subsubsection{Evaluation Metrics}

We evaluate detection performance using mAP, mAP$_{50}$, mAP$_{75}$, and Recall. mAP is computed by averaging precision across multiple IoU thresholds from 0.5 to 0.95, while mAP$_{50}$ and mAP$_{75}$ reflect coarse and strict localization performance, respectively. Average Recall measures the maximum recall under a fixed number of detections per image. To assess communication efficiency, we additionally report Input Bits Per Pixel (BPP), which quantifies the effective number of transmitted bits normalized by the original image size. Depending on the pipeline, Input BPP is calculated from the total size of the compressed image for compressed-image baselines, or from the number of selected tokens, feature dimensionality, and position information for token-based pipelines, with or without vector quantization. This measure captures the full transmission payload, including both feature representation and spatial location, enabling a direct comparison of accuracy versus communication cost across different approaches.
\begin{figure*}[t]
  \centering
  \includegraphics[width=0.8\textwidth]{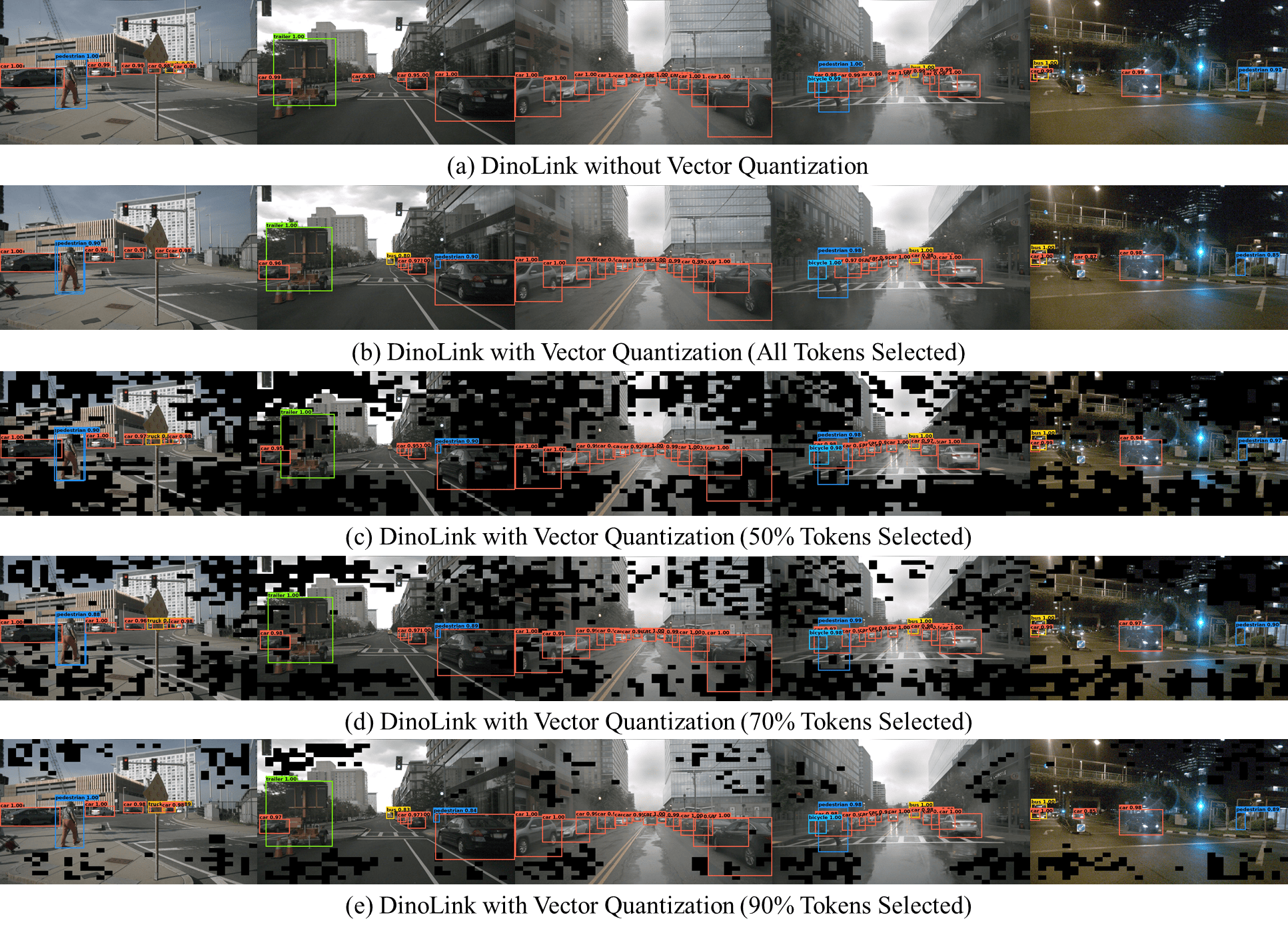}
  \vspace{-15pt}
  \caption{Qualitative detection results under different token ratios. Top rows are full-frame baselines; bottom rows keep only selected DINO patches. Higher token ratios (50\%→70\%→90\%) preserve more details and improve detection quality.}
  \label{fig:detection_visualization}
  \vspace{-15pt}
\end{figure*}

\subsection{Main Experiments and Results}

Table \ref{table:baseline_comparison} compares our framework against raw input and standard codecs (JPEG, WebP).

\subsubsection{Transmission Efficiency} Our VQ-Main achieves an extreme compression of 0.021 BPP, representing a 146$\times$ bitrate reduction compared to the uncompressed baseline (2.92 BPP). While traditional WebP (Q80) still requires 22$\times$ more bandwidth (0.45 BPP), our method maintains a competitive 32.8\% mAP, proving its superior efficiency for bandwidth-constrained V2X communication.

\subsubsection{Task-Aware Robustness} 
The results indicate that the gap in detection accuracy is remarkably narrow considering the orders-of-magnitude difference in bitrates. Unlike human-centric codecs (JPEG/WebP) that prioritize visual smoothness, our framework focuses on the structural integrity of objects. Consequently, our approach achieves a much higher "information density" per bit for the detection task, proving that pixel-perfect reconstruction is not a prerequisite for high-performance V2X perception.

\begin{table}[t]
\caption{Comparison with Standard Compression Methods}
\vspace{-10pt}
\label{table:baseline_comparison}
\centering
\resizebox{\columnwidth}{!}{
\begin{tabular}{lccccc}
\toprule
Method & BPP $\downarrow$ & mAP $\uparrow$ & mAP$_{50}$ $\uparrow$ & mAP$_{75}$ $\uparrow$ & Recall $\uparrow$ \\
\midrule

JPEG (Q100) & 2.641& \textbf{38.8\%} & 69.2\% & 36.2\% & \textbf{48.7\%} \\
JPEG (Q90)  & 1.140& \textbf{38.8\%} & \textbf{69.3\%} & 35.8\% & \textbf{48.7\%} \\
WebP (Q90)  & 0.816& 38.7\% & 68.8\% & 36.0\% & 48.5\% \\
WebP (Q80)  & 0.452& 38.1\% & 69.0\% & \textbf{36.7\%} & 48.3\% \\
\midrule

Dino (no Comp.) & 2.920& \textbf{38.8\%} & 69.2\% & 36.3\% & 48.6\% \\
\midrule

\textbf{Ours (with RVQ)} & \textbf{0.021}& 32.8\% & 63.0\% & 28.6\% & 43.6\% \\
\bottomrule
\end{tabular}
} 
\vspace{-20pt}
\end{table}

\subsection{Impact of Top-$K$ Token Selection Ratio}

\begin{figure}[t] 
  \centering
  \includegraphics[width=0.8\columnwidth]{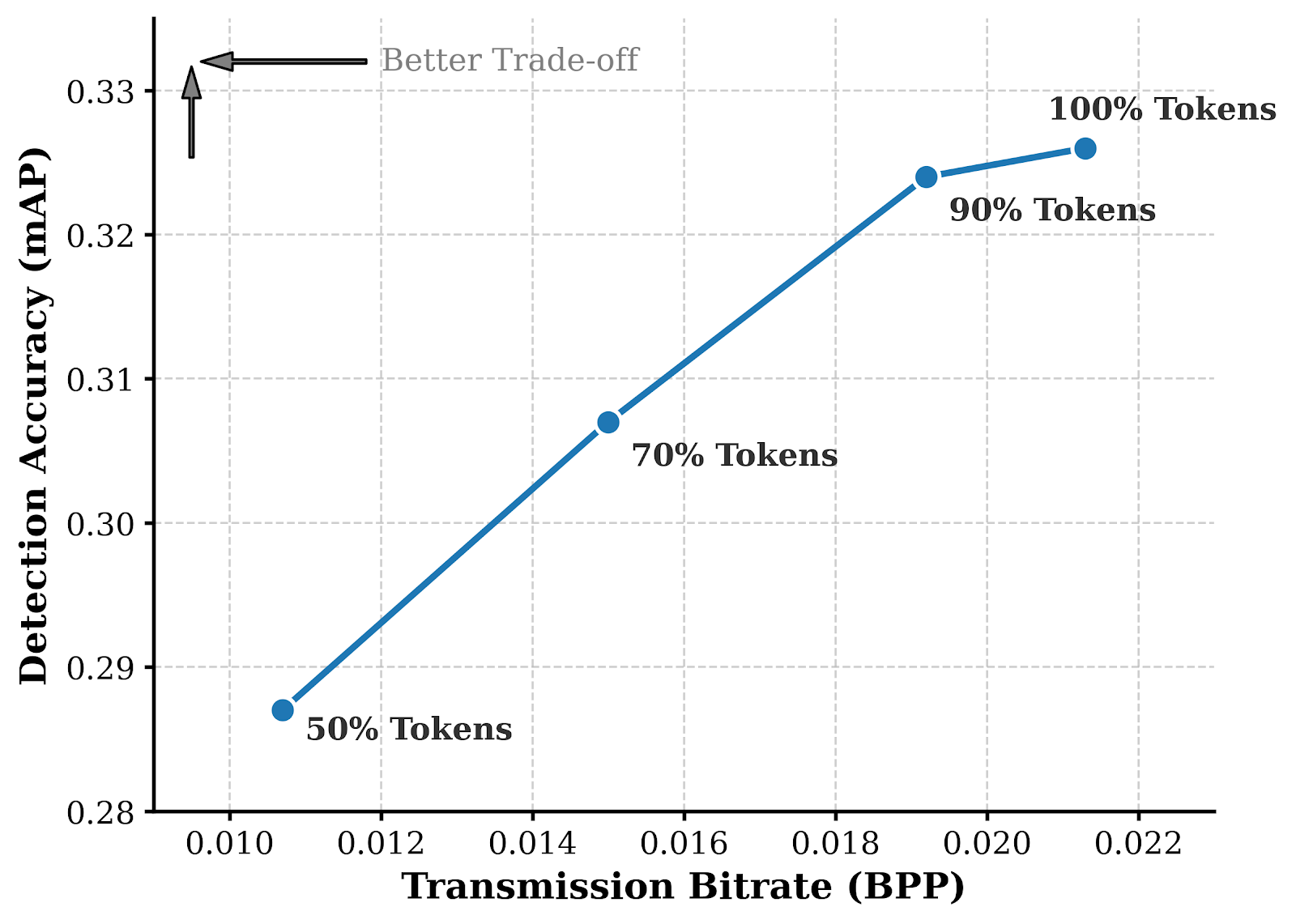}
  \vspace{-15pt}
  \caption{Efficiency-accuracy trade-off analysis. The chart illustrates the impact of Top-$K$ token selection ratios on transmission bitrate (BPP) and detection mAP.}
  \label{fig:pareto_chart}
  \vspace{-18pt}
\end{figure}

To evaluate the impact of task-aware token pruning, we visualize the detection results under various selection ratios in Fig. \ref{fig:detection_visualization}. Even as the selection ratio decreases to 50\% and significant background regions are masked, the framework consistently prioritizes and preserves tokens corresponding to critical objects such as vehicles and pedestrians. The model maintains stable bounding box predictions and high confidence scores across diverse urban scenes, confirming that our Top-$K$ strategy effectively filters out environmental noise without compromising the structural integrity required for robust remote perception.This qualitative resilience is further substantiated by the quantitative results illustrated in Fig. \ref{fig:pareto_chart} and detailed in Table \ref{table:topk_ablation}. As the selection ratio decreases from 100\% to 50\%, we observe a near-linear reduction in transmission BPP from 0.021 to 0.011. However, the detection performance exhibits a resilient non-linear degradation. Notably, reducing the ratio to 90\% achieves a BPP of 0.019 while maintaining a competitive 32.4\% mAP, representing only a marginal 0.2\% absolute drop compared to the full-token baseline (32.6\% mAP). These findings suggest significant redundancy in visual tokens for autonomous driving scenarios, where a subset of task-relevant tokens is sufficient to preserve essential semantic information. Consequently, the 90\% ratio represents a Pareto-optimal configuration, successfully maximizing communication efficiency while preserving the requisite accuracy for high-fidelity detection.

\begin{table}[t]
\caption{Impact of Top-$K$ selection ratio on transmission bitrate and detection performance ($Codebook\ Size = 768$). }
\vspace{-10pt}
\label{table:topk_ablation}
\centering
\resizebox{\columnwidth}{!}{
\begin{tabular}{lccccc}
\toprule
Ratio & BPP $\downarrow$ & mAP $\uparrow$ & mAP$_{50}$ $\uparrow$ & mAP$_{75}$ $\uparrow$ & Recall $\uparrow$ \\
\midrule
100\% (Full) & 0.021 & 32.8\%& 63.0\%& 28.6\%& 43.6\%\\
\textbf{90\%} & \textbf{0.019} & \textbf{32.4\%} & \textbf{62.2\%} & \textbf{28.1\%} & \textbf{43.3\%} \\
70\% & 0.015 & 30.7\% & 59.5\% & 27.6\% & 41.7\% \\
50\% & 0.011 & 28.7\% & 57.8\% & 24.7\% & 39.0\% \\
\bottomrule
\end{tabular}
}
\vspace{-10pt}
\end{table}

\subsection{Analysis of Quantization Capacity}

To evaluate the representational capacity of our framework, we investigate the impact of different codebook sizes within the RVQ module. As summarized in Table \ref{table:codebook_comparison}, we compare three configurations (512, 768, and 1024) to identify the optimal latent space density for urban driving scenes.

The results indicate that a codebook size of 768 serves as the optimal configuration, achieving the highest detection accuracy of 32.8\% mAP and a superior codebook utilization rate of 61.2\%. When the size is increased to 1024, we observe a performance degradation to 27.0\% mAP, accompanied by a significant drop in utilization to 38.6\%. This phenomenon, characterized by lower Perplexity (116.5 vs. 181.2), suggests that an over-parameterized discrete space leads to codebook collapse, where a large portion of the codebook remains inactive during inference. Conversely, while a smaller size of 512 maintains high utilization, its limited capacity results in insufficient expressive power to capture the complex semantic features of small objects, leading to a reduced mAP of 27.0\%. These findings confirm that the 768-entry configuration provides the ideal balance between quantization stability and feature reconstruction accuracy.

\begin{table}[t]
\caption{Comparison of Different Codebook Sizes in RVQ Module}
\vspace{-10pt}
\label{table:codebook_comparison}
\centering
\resizebox{\columnwidth}{!}{
\begin{tabular}{lccccccc}
\toprule
Size & BPP $\downarrow$ & Util. $\uparrow$ & Perp. $\uparrow$ & mAP $\uparrow$ & mAP$_{50}$ $\uparrow$ & mAP$_{75}$ $\uparrow$ & Recall $\uparrow$ \\
\midrule
512  & 0.019 & 60.4\% & 92.1 & 27.0\% & 54.3\% & 23.2\% & 37.5\% \\
\textbf{768} & \textbf{0.021} & \textbf{61.2\%} & \textbf{181.2} & \textbf{32.8\%} & \textbf{63.0\%} & \textbf{28.6\%} & \textbf{43.6\%} \\
1024 & 0.021 & 38.6\% & 116.5 & 27.0\% & 52.8\% & 24.3\% & 37.5\% \\
\bottomrule
\end{tabular}
}
\vspace{-20pt}
\end{table}

\subsection{Deployment Feasibility and Network Resilience}
\begin{figure}[t]
\vspace{-5pt}
  \centering
  \includegraphics[width=\columnwidth]{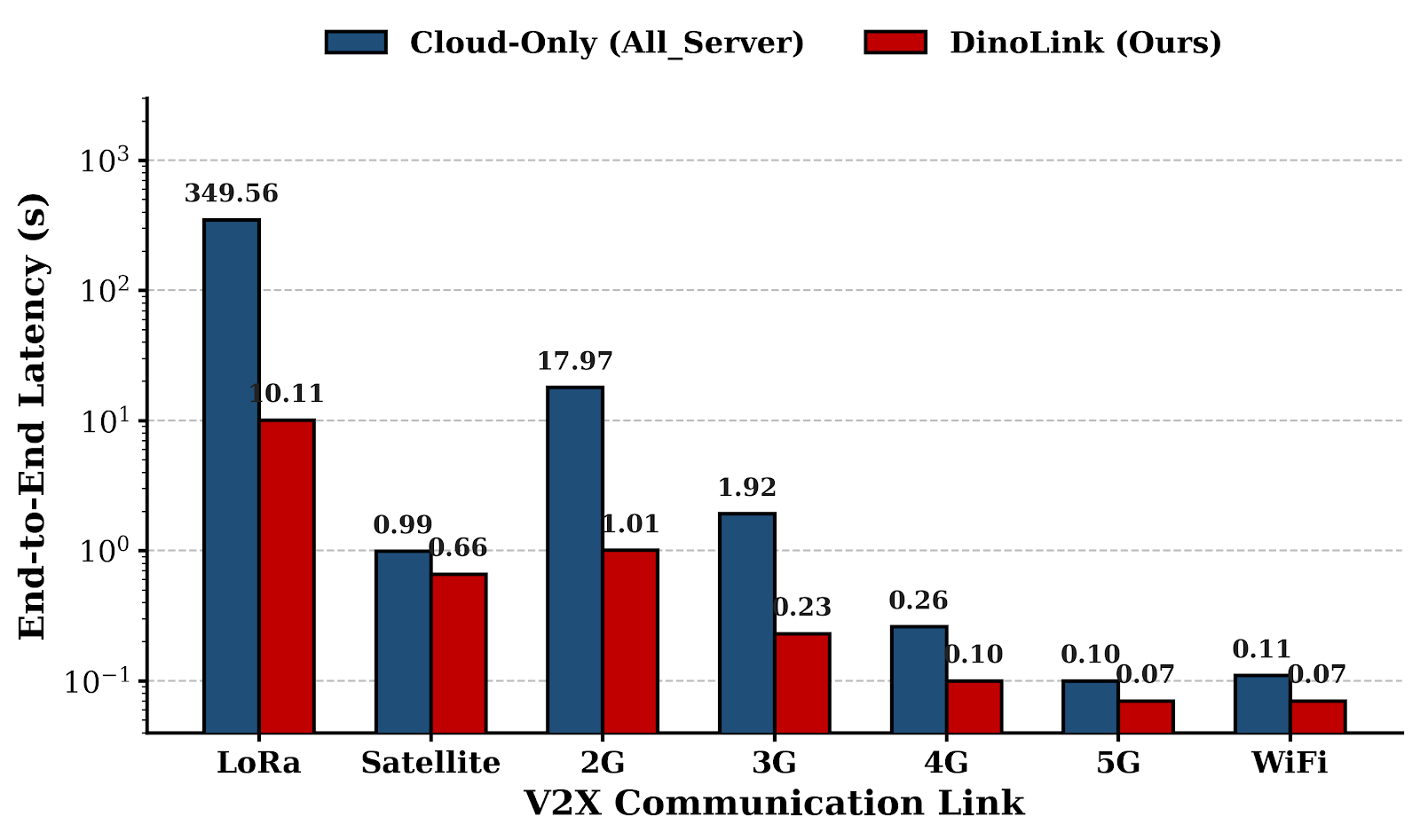}
  \vspace{-10pt}
  \caption{End-to-end latency comparison across diverse communication links. DinoLink demonstrates superior resilience in bandwidth-constrained environments (e.g., LoRa and 2G) by significantly reducing the transmission payload.}
  \label{fig:latency_comparison}
  \vspace{-20pt}
\end{figure}
To evaluate the practical utility of our framework, we simulated end-to-end latency across a spectrum of network conditions, ranging from narrow-band (LoRa, 2G) to high-speed (5G, WiFi) standards. We compared Two deployment paradigms: All-Server and the proposed DinoLink (Partition). We do not present an All-Local execution case as it is inherently impractical for the target scenarios of DinoLink. Our framework is specifically designed for resource-constrained edge devices tasked with computationally heavy models. While our experiments were conducted on a high-performance NVIDIA RTX 6000 Blackwell GPU for benchmarking, typical edge deployments rely on low-power embedded processors that lack the memory and TFLOPS required to run complex semantic encoders locally at viable frame rates. Under such hardware constraints, the latency of All-Local execution would be prohibitive, making the offloading of neural-compressed features to a robust server-side resource not just an optimization, but a necessity for real-time V2X perception.

As illustrated in Fig.~\ref{fig:latency_comparison}, DinoLink exhibits superior resilience to network degradation compared to traditional approaches. In extreme constraints like LoRa (0.005 Mbps), the All-Server mode becomes prohibitive with a latency of 349.56s, whereas DinoLink maintains operability at 10.11s, achieving a $34.5\times$ acceleration. Similarly, in 2G networks, DinoLink (1.01s) outperforms All-Server (17.97s) by an order of magnitude. In high-speed 5G and WiFi environments, DinoLink achieves near-real-time performance of 0.07s, matching server-centric efficiency while benefiting from task-oriented feature pruning. These findings substantiate that our framework drastically lowers bandwidth requirements for remote perception. By transmitting quantized, task-relevant latent tokens rather than raw pixels, DinoLink ensures robust V2X performance even in wide-area or severely degraded network environments.

\subsection{Ablation Study}
We evaluate the impact of the RVQ module by comparing our framework against a baseline model operating without quantization. As summarized in Table \ref{table:vq_ablation}, the absence of VQ necessitates a prohibitive transmission bitrate of 2.92 BPP. In contrast, DinoLink achieves a $139\times$ bitrate reduction to 0.021 BPP. While the baseline naturally yields a superior detection accuracy of 38.8\% mAP due to the lack of information loss, DinoLink maintains a highly competitive 32.8\% mAP, offering a significantly more efficient trade-off for bandwidth-constrained environments. Qualitative results in Fig. \ref{fig:detection_visualization} substantiate these findings: subfigure (a) illustrates the peak detection performance of the uncompressed baseline, while subfigure (b) demonstrates that our quantized approach successfully preserves critical semantic features for accurate localization despite the extreme reduction in data volume.

\begin{table}[h]
\vspace{-10pt}
\caption{Impact of RVQ on Bitrate and Accuracy}
\vspace{-5pt}
\label{table:vq_ablation}
\centering
\resizebox{\columnwidth}{!}{
\begin{tabular}{lccccc}
\toprule
Method & BPP $\downarrow$ & mAP $\uparrow$ & mAP$_{50}$ $\uparrow$ & mAP$_{75}$ $\uparrow$ & AR $\uparrow$ \\
\midrule
Baseline (w/o VQ) & 2.920 & 38.8\% & 69.2\% & 36.3\% & 48.6\% \\
DinoLink (w/ VQ) & 0.021 & 32.8\% & 63.0\% & 28.6\% & 43.6\% \\
\bottomrule
\end{tabular}
}
\vspace{-15pt}
\end{table}

\subsection{Real Experiment}

\begin{figure}[h]
  \centering
  \vspace{-15pt}
  \includegraphics[width=0.7\columnwidth]{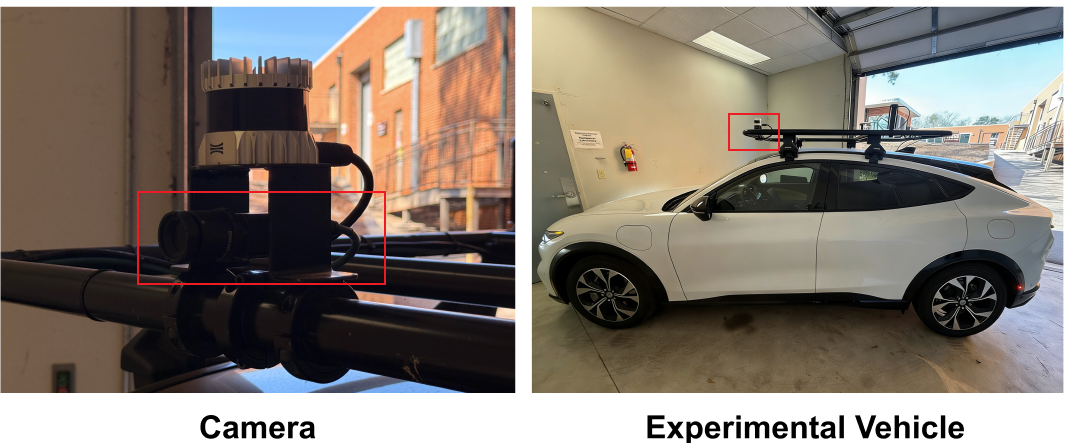}
  \caption{\textbf{Real-world experimental setup.} (Left) Roof-mounted camera for real-time visual capture at the vehicle edge. (Right) The experimental vehicle platform used to validate DinoLink in physical V2X scenarios.}
\label{fig:real_experiment}
\vspace{-10pt}
\end{figure}

\begin{figure}[h]
  \centering
  \includegraphics[width=0.7\linewidth]{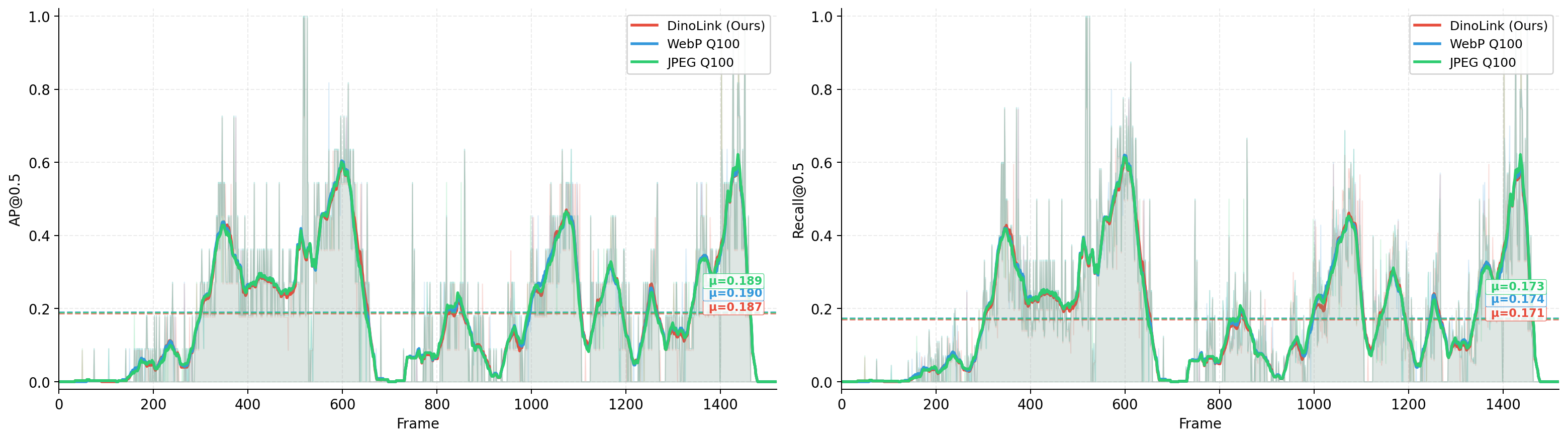}
  \vspace{-15pt}
  \caption{Per-frame detection metrics from our live vehicle-to-PC deployment ($K{=}70$).}
  \label{fig:perframe}
  \vspace{-15pt}
\end{figure}

We validate DinoLink via a live vehicle-to-server deployment over a Local Area Network (LAN), where a personal computer operates as the receiving edge server. Instead of relying on offline dataset replays, we process 1{,}520 frames (2448$\times$2048) streaming in real time from a roof-mounted camera on a physical vehicle (Fig.~\ref{fig:real_experiment}). Using uncompressed detections as pseudo-ground-truth, all methods are evaluated at a 0.3 confidence threshold.
As shown in Fig.~\ref{fig:perframe}, DinoLink ($K{=}70$) yields a mean AP@0.5 of 0.187, closely matching the WebP (0.190) and JPEG (0.189) Q100 baselines. Crucially, DinoLink operates at just 0.005 BPP---over 500$\times$ and 1{,}000$\times$ lower bandwidth than WebP and JPEG, respectively. This live physical deployment confirms that DinoLink effectively replaces traditional multimedia codecs for V2X perception, slashing the transmitted payload to ${\approx}$0.26~KB per frame with negligible accuracy loss.transmitted payload to ${\approx}$0.26~KB per frame with negligible accuracy loss.

\section{CONCLUSION}\label{sec:conclusion}In this paper, we presented \emph{DinoLink}, a token-centric transmission framework designed to bridge the gap between high-performance foundation models and bandwidth-constrained V2X perception. By integrating saliency-aware Top-$K$ selection with RVQ, DinoLink achieves a dual-sparsity funnel that aggressively prunes environmental redundancy and collapses continuous features into compact discrete indices. Our experimental results on the nuScenes dataset demonstrate that DinoLink facilitates a $139\times$ reduction in bitrate compared to uncompressed feature transmission, while maintaining a competitive 32.8\% mAP for downstream object detection. Furthermore, deployment simulations on high-performance NVIDIA RTX 6000 Blackwell hardware substantiate that our framework provides superior network resilience, achieving a $34.5\times$ latency acceleration under extreme narrow-band conditions compared to raw data offloading. By decoupling semantic representation from specific downstream architectures, DinoLink offers a scalable and efficient solution for robust vehicle-cloud collaborative perception in real-world autonomous driving scenarios.
\subsubsection{Limitations}While DinoLink demonstrates high efficiency, our current token selection strategy is primarily coupled with the DINOv2 self-attention mechanism, and the generalizability of using other saliency-based methods remains to be explored. Furthermore, the evaluation was conducted on a relatively small-scale dataset, necessitating further validation on larger autonomous driving benchmarks to ensure the robustness of our RVQ compression and Top 50\% selection strategy under more diverse environments.
\subsubsection{Future Work}Our future research will focus on extending the framework to support multi-vehicle fusion, enabling true cooperative perception by aggregating compressed token streams from multiple perspectives to overcome occlusions. Additionally, we aim to develop a dynamic adaptation mechanism that can automatically adjust the token selection ratio and RVQ parameters based on real-time V2X bandwidth fluctuations to maintain an optimal balance between perception accuracy and latency.

\bibliographystyle{IEEEtran}
\bibliography{ref}

\end{document}